\documentclass[10pt,twocolumn,letterpaper]{article}
\usepackage{cvpr}
\usepackage{times}
\usepackage{graphicx}
\usepackage{amsmath}
\usepackage{amssymb}
\usepackage{xspace}
\usepackage{xcolor}
\usepackage{enumitem}
\usepackage{listings}
\usepackage{tabularx}
\usepackage{multirow}
\usepackage[export]{adjustbox}
\usepackage{tikz}
\usepackage{pgffor}
\usepackage[numbers,square,sort&compress]{natbib}
\usepackage[
  pagebackref=true,
  breaklinks=true,
  letterpaper=true,
  colorlinks,
  bookmarks=false]{hyperref}

\cvprfinalcopy
\ifcvprfinal\fi

\graphicspath{{./figures}{./}{./figures/v14}}

\setitemize{noitemsep,topsep=0pt,parsep=0pt,partopsep=0pt,nolistsep,leftmargin=*}

\renewcommand{\paragraph}[1]{\medskip\noindent{\bf #1}}

\definecolor{attr}{rgb}{0.85,0.88,0.90}
\definecolor{code}{rgb}{0.95,0.93,0.90}

\newcommand{\off}[1]{}

\newcommand{\dcnn}{FV-CNN\xspace}
\newcommand{\rcnn}{FC-CNN\xspace}

\newcommand{\ktb}{KTH-T2b\xspace}

\newcommand{\pmt}[1]{{\tiny $\pm$#1}}

\newcolumntype{H}{>{\setbox0=\hbox\bgroup}c<{\egroup}@{}}

\makeatletter
\g@addto@macro\normalsize{%
  \setlength\abovedisplayskip{0.7em}
  \setlength\belowdisplayskip{0.7em}
  \setlength\abovedisplayshortskip{0.7em}
  \setlength\belowdisplayshortskip{0.7em}
}
\makeatother

\tikzset{
 image label/.style={
   fill=white,
   text=black,
   font=\footnotesize,
   anchor=south east,
   xshift=-0.1cm,
   yshift=0.1cm,
   at={(0,0)}
 }
}

\begin{document}
\title{Deep Filter Banks for Texture Recognition and Segmentation}

\vspace{-2em}
\author{
Mircea Cimpoi\\
University of Oxford \\
\small \url{mircea@robots.ox.ac.uk}
\and
Subhransu Maji\\
University of Massachusetts, Amherst \\
\small \url{smaji@cs.umass.edu}
\and
Andrea Vedaldi \\
University of Oxford \\
\small \url{vedaldi@robots.ox.ac.uk} \\
}
\maketitle

\begin{abstract}
Research in texture recognition often concentrates on the problem of material recognition in uncluttered conditions, an assumption rarely met by applications. In this work we conduct a first study of material and describable texture attributes recognition in clutter, using a new dataset derived from the OpenSurface texture repository. Motivated by the challenge posed by this problem, we propose a new texture descriptor, \dcnn, obtained by Fisher Vector pooling of a Convolutional Neural Network (CNN) filter bank. \dcnn substantially improves the state-of-the-art in texture, material and scene recognition. Our approach achieves 79.8\% accuracy on Flickr material dataset and 81\% accuracy on MIT indoor scenes, providing absolute gains of more than 10\% over existing approaches. \dcnn easily transfers across domains without requiring feature adaptation as for methods that build on the fully-connected layers of CNNs. Furthermore, \dcnn can seamlessly incorporate multi-scale information and describe regions of arbitrary shapes and sizes. Our approach is particularly suited at localizing ``stuff'' categories and obtains state-of-the-art results on MSRC segmentation dataset, as well as promising results on recognizing materials and surface attributes in clutter on the OpenSurfaces dataset.
\end{abstract}

\section{Introduction}\label{s:intro}
Texture is ubiquitous and provides useful cues of material properties of objects and their identity, especially when shape is not useful. Hence, a significant amount of effort in the computer vision community has gone into recognizing texture via tasks such as texture perception~\cite{adelson01on-seeing,amadasun89textural,gaarding1992shape,forsyth2001shape} and description~\cite{ferrari07learning,cimpoi14describing}, material recognition~\cite{leung2001representing,sharan13recognizing,schwartz13visual}, segmentation~\cite{manjunath1991unsupervised,jain1991unsupervised}, and even synthesis~\cite{efros1999texture,wei2000fast}.

Perhaps the most studied task in texture understanding is the one of material recognition, as captured in benchmarks such as CuRET~\cite{dana99reflectance}, KTH-TIPS~\cite{caputo05class}, and, more recently, FMD~\cite{sharan09material}. However, while at least the FMD dataset contains images collected from the Internet, vividly dubbed ``images in the wild'', all these datasets make the simplifying assumption that textures fill images. Thus, they are not necessarily representative of the significantly harder problem of recognising materials in natural images, where textures appear in clutter. Building on a recent dataset collected by the computer graphics community, the {\bf first contribution} of this paper is a {\em large-scale analysis of material and perceptual texture attribute recognition and segmentation in clutter} (Fig.~\ref{fig:first-page} and Sect.~\ref{s:data}).

Motivated by the challenge posed by recognising texture in clutter, we develop a new texture descriptor.  In the simplest terms a texture is characterized by the arrangement of local patterns, as captured in early works~\cite{leung2001representing,varma2003texture} by the distribution of local ``filter bank'' responses. These filter banks were designed to capture edges, spots and bars at different scales and orientations. Typical combinations of the filter responses, identified by vector quantisation, were used as the computational basis of the ``textons" proposed by Julesz~\cite{julesz83textons}. Texton distributions were the early versions of ``bag-of-words'' representations, a dominant approach in recognition in the early 2000s, since then improved by new pooling schemes such as soft-assignment~\cite{wang2010locality,zhou2010image,liu2011defense} and Fisher Vectors (FVs)~\cite{perronnin2010improving}. Until recently, FV with SIFT features~\cite{lowe99object} as a local representation was the state-of-the-art method for recognition, not only for textures, but for objects and scenes too.

\newcommand{\puti}[2]
{%
\begin{tikzpicture}
\node[anchor=south east,inner sep=0] at (0,0) {#2};
\node[image label]{#1};
\end{tikzpicture}%
}

\begin{figure}
\puti{banded}{\includegraphics[width=0.115\textwidth,height=0.1\textwidth]{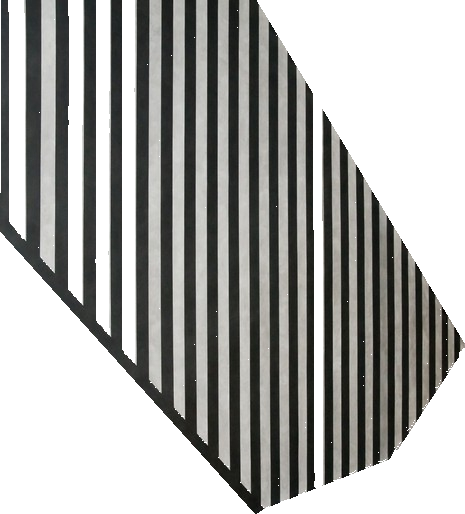}}
\puti{blotchy}{\includegraphics[width=0.115\textwidth,height=0.1\textwidth]{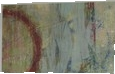}}
\puti{chequered}{\includegraphics[width=0.115\textwidth,height=0.1\textwidth]{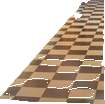}}
\puti{grid}{\includegraphics[width=0.115\textwidth,height=0.1\textwidth]{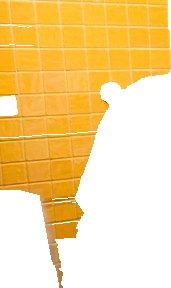}}
\\
\puti{marbled}{\includegraphics[width=0.115\textwidth,height=0.1\textwidth]{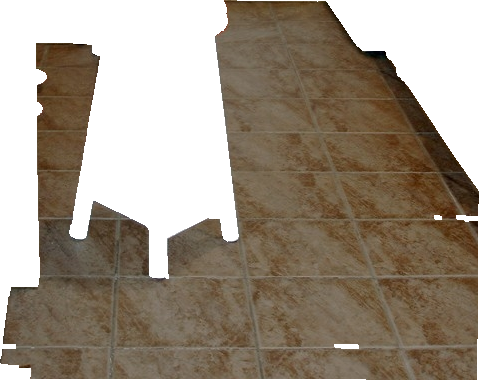}}
\puti{paisley}{\includegraphics[width=0.115\textwidth,height=0.1\textwidth]{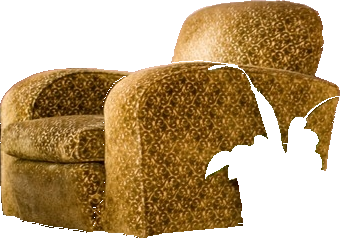}}
\puti{paisley}{\includegraphics[width=0.115\textwidth,height=0.1\textwidth]{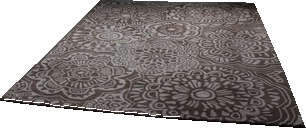}}
\puti{wrinkled}{\includegraphics[width=0.115\textwidth,height=0.1\textwidth]{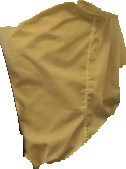}}
\\
\puti{brick}{\includegraphics[width=0.115\textwidth,height=0.1\textwidth]{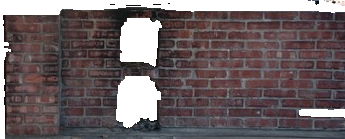}}
\puti{ceramic}{\includegraphics[width=0.115\textwidth,height=0.1\textwidth]{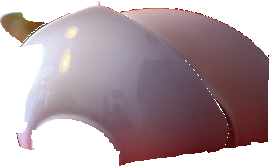}}
\puti{carpet}{\includegraphics[width=0.115\textwidth,height=0.1\textwidth]{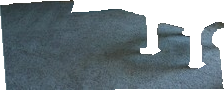}}
\puti{fabric}{\includegraphics[width=0.115\textwidth,height=0.1\textwidth]{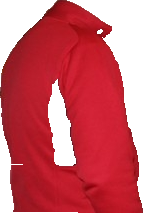}}
\caption{{\bf Texture recognition in clutter}. Example of top retrieved texture segments by attributes (top two rows) and materials (bottom) in the OpenSurfaces dataset.}
 \label{fig:first-page}
\end{figure}

Later, however, Convolutional Neural Networks (CNNs) have emerged as the new state-of-the-art for recognition, exemplified by remarkable results in image classification~\cite{krizhevsky12imagenet}, detection~\cite{girshick14rich} and segmentation~\cite{hariharan2014simultaneous} on a number of widely used benchmarks. Key to their success is the ability to leverage large \emph{labelled} datasets to learn increasingly complex transformations of the input to capture invariances. Importantly, CNNs pre-trained on such large datasets have been shown~\cite{oquab14learning,chatfield14return,girshick14rich} to contain general-purpose feature extractors, transferrable to many other domains.

Domain transfer in CNNs is usually achieved by using as features the output of a deep, fully-connected layer of the network. From the perspective of textures, however, this choice has three drawbacks. The first one (I) is that, while the convolutional layers are akin to non-linear filter banks, the fully connected layers capture their spatial layout. While this may be useful for representing the shape of an object, it may not be as useful for representing texture. A second drawback (II) is that the input to the CNN has to be of  fixed size to be compatible with the fully connected layers, which requires an expensive resizing of the input image, particularly when features are computed for many different regions~\cite{girshick14rich,gong14multi-scale}. A third and more subtle drawback (III) is that deeper layers may be more domain-specific and therefore potentially less transferrable than shallower layers.


The {\bf second contribution} of this paper is \dcnn (Sect.~\ref{s:method}), a {\em pooling method that overcomes these drawbacks}. The idea is to regard the convolutional layers of a CNN as a filter bank and build an orderless representation using FV as a pooling mechanism, as is commonly done in the bag-of-words approaches. Although the suggested change is simple, the approach is remarkably flexible and effective. First, pooling is orderless and multi-scale, hence suitable for textures. Second, any image size can be processed by convolutional layers, avoiding costly resizing operations. Third, convolutional filters, pooled by \dcnn, are shown to transfer more easily than fully-connected ones even without fine-tuning. While other authors \cite{he14spatial,gong14multi-scale} have recently proposed alternative pooling strategies for CNNs, we show that our method is more natural, faster and often significantly more accurate.

The {\bf third contribution} of the paper is a {\em thorough evaluation of these descriptors} on a variety of benchmarks, from textures to objects (Sect.~\ref{s:results}). In textures, we evaluate material and describable attributes recognition and segmentation on new datasets derived from OpenSurfaces (Sect.~\ref{s:data}). When used with linear SVMs, \dcnn improves the state of the art on texture recognition by a significant margin.
Like textures, scenes are also weakly structured and a bag-of-words representation is effective. \dcnn obtains 81.1\% accuracy on the MIT indoor scenes dataset~\cite{quattoni09recognizing}, significantly outperforming the current state-of-the-art of 70.8\%~\cite{zhou14learning}. What is remarkable is that, where~\cite{zhou14learning} finds that CNNs trained on scene recognition data perform better than CNNs trained on an object domain (ImageNet), when used in \dcnn not only is there an overall performance improvement, but the domain-specific advantage is entirely removed (Tab.~\ref{t:places-cnn}). This indicates that \dcnn are in fact better at domain transfer. Our method also matches the previous best in PASCAL VOC 2007 classification dataset providing measurable boost over CNNs and closely approaches competitor methods on CUB 2010-2011 datasets when ground-truth object bounding boxes are given.



\dcnn can be used for describing regions by simply pooling across pixels within the region. Combined with a low-level segmentation algorithm this suffices to localize textures within images. This approach is similar to a recently proposed method called ``R-CNN" for localizing objects~\cite{girshick14rich}. However, in contrast to it we do not need repeated evaluations of the CNN since the convolutional features can be computed just once for the entire image and pooled differently. This makes \dcnn not only faster, but also as experiments suggest, much more accurate at texture localization. We achieve state of the art results on the MSRC segmentation dataset using a simple scheme of classifying ``superpixels" obtaining an accuracy of 87.0\% (previous best 86.5\%). The corresponding R-CNN obtains 57.7\% accuracy. Segmentation results are promising in the \emph{OpenSurfaces} dataset~\cite{bell13opensurfaces} as well.

Finally, we analyze the utility of different network layers and architectures as filter banks, concluding that: SIFT is competitive only with the first few layers of a CNN (Fig.~\ref{f:filterbank}) and that significant improvement to the underlying CNN architecture, such as the ones achieved by the \emph{very deep} models of Simonyan and Zisserman~\cite{simonyan14very}, directly translate into much better filter banks for texture recognition.

\section{Texture recognition in clutter}\label{s:data}

A contribution of this work is the analysis of materials and texture attributes in realistic imaging conditions. Earlier datasets such as KTH-TIPS were collected in controlled conditions, which makes their applicability to natural images unclear. More recent datasets such as FMD and DTD remove this limitation by building on images downloaded from the Internet, dubbed images ``in the wild''. However, in these datasets texture always fill the field of view of the camera. In this paper we remove this limitation by experimenting for the first time with a large dataset of textures collected in the wild and in cluttered conditions.

In particular, we build on the \emph{Open Surfaces} (OS) dataset that was recently introduced by Bell~\etal~\cite{bell13opensurfaces} in computer graphics. OS comprises 25,357 images, each containing a number of high-quality texture/material segments. Many of these segments are annotated with  additional attributes such as the material name, the viewpoint, the BRDF, and the object class. Not all segments have a complete set of annotations; the experiments in this paper focus on the 58,928 that contain material names. Since material classes are highly unbalanced, only the materials that contain at least 400 examples are considered. This result in 53,915 annotated material segments in 10,422 images spanning 22 different classes.\footnote{The classes and corresponding number of example segments are:
          brick (610),
      cardboard  (423),
     carpet/rug (1,975),
        ceramic (1,643),
       concrete  (567),
   fabric/cloth (7,484),
           food (1,461),
          glass (4,571),
 granite/marble (1,596),
           hair  (443),
          other (2,035),
       laminate  (510),
        leather  (957),
          metal (4,941),
        painted (7,870),
   paper/tissue (1,226),
  plastic/clear  (586),
 plastic/opaque (1,800),
          stone  (417),
           tile (3,085),
      wallpaper  (483),
           wood (9,232).}
Images are split evenly into training, validation, and test subsets with 3,474 images each. Segment sizes are highly variable, with half of them being relatively small, with an area smaller than $64 \times 64$ pixels. While the lack of exhaustive annotations makes it impossible to define a complete background class, several less common materials (including for example segments that annotators could not assign to a material) are merged in an ``other'' class that acts as pseudo-background.

In order to study perceptual properties as well as materials, we augment the OS dataset with some of the 47 attributes from the DTD~\cite{cimpoi14describing}. Since the OS segments do not trigger with sufficient frequency all the 47 attributes, the evaluation is restricted to eleven of them for which it was possible to identify at least 100 matching segments.\footnote{These are: banded, blotchy, chequered, flecked, gauzy, grid, marbled, paisley, pleated, stratified, wrinkled.} The attributes were manually labelled in the 53,915 segments retained for materials. We refer to this data as OSA. The complete list of images, segments, labels, and splits are available at \url{http://www.robots.ox.ac.uk/~vgg/data/wildtex/}.

\section{Method}\label{s:method}

This section describes the methodological contributions of this paper: region description and segmentation.

\subsection{Region description}\label{s:tex-desc}

This section introduces a number of visual descriptors suitable to model the appearance of image regions. Texture is traditionally described by orderless pooling of filter bank responses as, unlike in objects, the overall shape information is usually unimportant. However, small under-sampled textures may benefit if recognized in the context of an object. Thus, the primacy of orderless pooling may not always hold in the recognition of textures in natural conditions.

In order to explore the interplay between shape and orderless pooling, we evaluate two corresponding region descriptors: \rcnn for shape and \dcnn for texture. Both descriptors are based on the same CNN features~\cite{krizhevsky12imagenet} obtained from an off-the-shelf CNN pre-trained on the ImageNet ILSVRC 2012 data as suggested in~\cite{jia13caffe,razavin14cnn-features,chatfield14return}. Since the underlying CNN is the same, it is meaningful to compare FC- and \dcnn directly.

\paragraph{Object descriptor: \rcnn.} The \rcnn descriptor is obtained by extracting as features the output of the penultimate Fully-Connected (FC) layer of a CNN, including the non-linear gating function, applied to the input image. This can be considered an object descriptor because the fully connected layers allow \rcnn to \emph{capture the overall shape of the object} contained in the region. \rcnn is applied to an image region $R$ (which may be the whole image) by warping the bounding box enclosing  $R$ (plus a 10\% border) to a square of a fixed size matching the default CNN input geometry,  obtaining the same R-CNN descriptor introduced by Girshick~\etal~\cite{girshick14rich} as a state-of-the-art object detector in the PASCAL VOC~\cite{everingham07pascal} data.

\paragraph{Texture descriptor: \dcnn.} The \dcnn descriptor is inspired by the state-of-the-art texture descriptors of~\cite{cimpoi14describing} based on the Fisher Vector (FV). Differently from \rcnn, FV pools local features densely within the described regions \emph{removing global spatial information}, and is therefore more apt at describing textures than objects. Here FV is computed on the output of a single (last) convolutional layer of the CNN, but we compared features from other layers as well (Sect~\ref{s:exp-filterbank}). By avoiding the computation of the fully connected layers, the input image does not need to be rescaled to a specific size; in fact, the dense convolutional features are extracted at multiple scales and pooled into a single FV just like for SIFT. The pooled convolutional features are extracted immediately after the last linear filtering operator and are not otherwise normalised.

The \dcnn descriptor is related to the one proposed by Gong~\etal~\cite{gong14multi-scale}. There VLAD pooling, which is similar to FV, is applied to \rcnn-like descriptors extracted from densely sampled image windows. A key difference of \dcnn is that dense features are extracted from the convolutional rather than fully-connected layers. This is more natural, significantly more efficient (as it does not require recomputing the network for each extracted local descriptor) and, as shown in Sect.~\ref{s:results}, more accurate.

\subsection{Region segmentation}\label{s:tex-seg}

This section discusses our method to automatically partition an image into a number of recognisable regions. Inspired by Cimpoi~\etal~\cite{cimpoi14describing} that successfully ported object description methods to texture descriptors, here we propose a segmentation technique inspired by object detection. An increasingly popular method for object detection, followed for example by R-CNN~\cite{girshick14rich}, is to first propose a number of candidate object regions using low-level image cues, and then verifying a shortlist of such regions using a powerful classifier. Applied to textures, this requires a low-level mechanism to generate textured region proposals, followed by a region classifier. A key advantage of this approach is that it allows applying object- (\rcnn) and texture-like (\dcnn) descriptors alike. After proposal classification, each pixel can be assigned more than one label; this is solved with simple voting schemes, also inspired by object detections methods.


The paper explores two such region generation methods: the crisp regions of~\cite{isola14crisp} and the multi-scale combinatorial grouping of~\cite{arbelaez2014multiscale}. In both cases, region proposals are generated using low-level image cues, such as colour or texture consistency, as specified by the original methods. It would of course be possible to incorporate \rcnn and \dcnn among these energy terms to potentially strengthen the region generation mechanism itself. However, this contradicts partially the logic of the scheme, which breaks down the problem into cheaply generating tentative segmentations and then verifying them using a more powerful (and likely expensive) model. Furthermore, and more importantly, these cues focus on separating texture \emph{instances}, as presented in each particular image, whereas \rcnn and \dcnn are meant to identify a texture class. It is reasonable to expect instance-specific cues (say the colour of a painted wall) to be better for segmentation.

\section{Results}\label{s:results}

\begin{table*}[ht]
\setlength{\tabcolsep}{2pt}
\centering
\begin{tabular}{|cl|c|c|HHHccc|ccc|c|c|}
\hline
\multicolumn{2}{|c|}{dataset} &  meas. & \multirow{2}{*}{IFV} & \multicolumn{3}{H}{AlexNet} & \multicolumn{3}{c|}{VGG-M} & \multicolumn{3}{c|}{VGG-VD} & FV-SIFT & \multirow{2}{*}{SoA} \\
\multicolumn{2}{|c|}{} &              (\%)  &                    & FC            & FV                 & FC+FV              & FC            & FV                 & FC+FV              & FC            & FV                 & FC+FV              & FC+FV-VD           & \\

\hline
\hline
\multirow{3}{*}{(a)} & \ktb      & acc   & 70.8\pmt{2.7}      & 71.5\pmt{1.3} & 69.7\pmt{3.2}      & 72.1\pmt{2.8}      & 71\pmt{2.3}   & 73.3\pmt{4.7}      & 73.9\pmt{4.9}      & 75.4\pmt{1.5} & \bf{81.8\pmt{2.5}} & \bf{81.1\pmt{2.4}} & \bf{81.5\pmt{2.0}} & 76.0\pmt{2.9} \cite{sulc14fast}                          \\
                     & FMD        & acc   & 59.8\pmt{1.6}      & 64.8\pmt{1.8} & 67.7\pmt{1.5}      & 71.4\pmt{1.7}      & 70.3\pmt{1.8} & 73.5\pmt{2.0}      & 76.6\pmt{1.9}      & 77.4\pmt{1.8} & 79.8\pmt{1.8}      & \bf{82.4\pmt{1.5}} & \bf{82.2\pmt{1.4}} & 57.7\pmt{1.7} \cite{qi14pairwise, sharan13recognizing} \\
                     & OS+R       & acc   & 39.8               & 36.8          & 46.1               & 49.8                 & 41.3          & 52.5               & 54.9               & 43.4          & 59.5               & \bf{60.9}               & 58.7               & --                                                       \\
\hline
\multirow{2}{*}{(b)} & DTD        & acc   & 58.6\pmt{1.2}      & 55.1\pmt{0.6} & 62.9\pmt{1.4}      & 66.5\pmt{1.1}      & 58.8\pmt{0.8} & 66.8\pmt{1.6} & 69.8\pmt{1.1} & 62.9\pmt{0.8} & 72.3\pmt{1.0} & \bf{74.7\pmt{1.0}}  & \bf{75.5\pmt{0.8}}   & --                                                       \\

                     & OSA+R      & mAP   & 56.5               & 53.9          & 62.1               & 64.6               & 54.3          & 65.2               & \bf{67.9}               & 49.7          & 67.2               & \bf{67.9}               & \bf{68.2}               & --                                                       \\
\hline
\multirow{6}{*}{(d)} & MSRC+R     & acc   & 85.7               & 83.6          & 91.7               & 94.9               & 85.0          & 95.4               & 96.9               & 79.4          & 97.7               & \bf{98.8}               & \bf{99.1}               & --                                       \\
                     & MSRC+R     &msrc-acc& 92.0               & 84.1          & 95.0               & 97.3               & 84.0          & 97.6               & 98.1               & 82.0          & \bf{99.2}               & \bf{99.6}              & \bf{99.5}               & --                                       \\
                     & VOC07      & mAP11   & 59.9               & 74.0          & 73.1               & 76.8               & 76.8          & 76.4               & 79.5               & 81.7          & \bf{84.9}               & \bf{85.1}               & \bf{84.5}               & 85.2 \cite{wei14cnn:}                    \\
                     & MIT Indoor & acc   & 54.9               & 58.6          & 69.7               & 71.6               & 62.5          & 74.2               & 74.4               & 67.6          & \bf{81.0}               & \bf{80.3}               & 80.0               & 70.8 \cite{zhou14learning}               \\
                     & CUB        & acc   & 17.5               & 45.8          & 49                 & 54.1                & 46.1          & 49.9                & 54.9               & 54.6          & \bf{66.7}         & \bf{67.3}          & 65.4         & 73.9 (62.8$^*$) \cite{zhang14part-based} \\
                     & CUB+R      & acc   & 27.7               & 54.5          & 62.6               & 65.2               & 56.5          & 65.5               & 68.1                & 62.8          & 73.0               & \bf{74.9}               & 73.6               & 76.37 \cite{zhang14part-based}           \\
\hline
\end{tabular}
\caption{Evaluation of texture descriptors. The table compares \rcnn, \dcnn on two networks trained on ImageNet -- VGG-M and VGG-VD, and IFV on dense SIFT.
We evaluated these descriptors on (a) material datasets (FMD, \ktb, OS+R), (b) texture attributes (DTD, OSA+R) and (c) general categorisation datasets (MSRC+R,VOC07,MIT Indoor) and fine grained categorisation (CUB, CUB+R). For this experiment the region support is assumed to be known (and equal to the entire image for all the datasets except OS+R and MSRC+R and for CUB+R, where it is set to the bounding box of a bird). ($^*$) using a model without parts. Best results are marked in bold.}
\label{f:tex-baseline}
\end{table*}

\begin{table*}
\centering
\begin{tabular}{|cc|ccc|ccc|c|}
\hline
& & \multicolumn{3}{c|}{VGG-M}  & \multicolumn{3}{c|}{VGG-VD}  & \\
dataset  & measure (\%) & \rcnn & \dcnn       & FV+FC-CNN & \rcnn & \dcnn       & FC+FV-CNN & SoA                       \\
\hline
     OS  & pp-acc          & 36.3  & 48.7 (46.9) & 50.5      & 38.8  & \bf{55.4 (55.7)} & \bf{55.2}      & --                        \\
   MSRC  & msrc-acc      & 56.1  & 82.3        & 75.5      & 57.7  & \bf{87.0}        & 80.4      & 86.5~\cite{ladicky10graph} \\
\hline
\end{tabular}
\caption{Segmentation and recognition using crisp region proposals of materials (OS) and things \& stuff (MSRC). Per-pixel accuracies are reported, using the MSRC variant (see text) for the MSRC dataset. Results using MCG proposals~\cite{arbelaez2014multiscale} are reported in brackets for FV-CNN.} \label{f:seg-table}
\end{table*}

This section evaluates the proposed region recognition methods for classifying and segmenting materials, describable texture properties, and higher-level object categories. Sect.~\ref{s:exp-rec} evaluates the \emph{classification task} by assessing how well regions can be classified given that their true extent is known and Sect.~\ref{s:exp-seg} evaluates both \emph{classification and segmentation}. The rest of the section introduces the evaluation benchmarks and technical details of the representations.

\paragraph{Datasets.} The evaluation considers three texture recognition benchmarks other than OS (Sect.~\ref{s:data}). The first one is the \emph{Flickr Material Dataset} (FMD)~\cite{sharan13recognizing}, a recent benchmark containing 10 material classes. The second one is the \emph{Describable Texture Datasets} (DTD)~\cite{cimpoi14describing}, which contains texture images \emph{jointly} annotated with 47 describable attributes drawn from the psychological literature. Both FMD and DTD contain images ``in the wild'', \ie collected in uncontrolled conditions. However, differently from OS, these images are uncluttered. The third texture dataset is \emph{KTH-TIPS-2b}~\citep{caputo05class,hayman04learning}, containing a number of example images for each of four samples of 11 material categories. For each material, images of one sample are used for training and the remaining for testing.

Object categorisation is evaluated in the \emph{PASCAL VOC 2007}~\cite{everingham07pascal} dataset, containing 20 object categories, any combination of which may be associated to any of the benchmark images. Scene categorisation uses the \emph{MIT Indoor}~\cite{quattoni09recognizing} dataset, containing 67 indoor scene classes. Fine-grained categorisation uses the \emph{Caltech/UCSD Bird} dataset (CUB)~\cite{welinder10caltech-ucsd}, containing images of 200 bird species.

Note that some of these datasets come with ground truth region/object localisation. The +R suffix will be appended to a dataset to indicate that this information is used both at training and testing time. For example, OS means that segmentation is performed automatically at test time, whereas OS+R means that ground-truth segmentations are used.

\paragraph{Evaluation measures.} For each dataset the corresponding standard evaluator protocols and accuracy measures are used. In particular, for FMD, DTD, MIT Indoor, CUB, and OS+R, evaluation uses average classification accuracy, per-segment/image and normalized for each class. When evaluating the quality of a segmentation algorithm, however, one must account for the fact that in most datasets, and in OS and MSRC in particular, not all pixels are labelled. In this case, accuracy is measured per-pixel rather than per-segment, ignoring all pixels that are unlabelled in the ground truth.
For MSRC, furthermore, accuracy is normalised across all pixels regardless of their category. For OSA, since some segments may have more than one label, we are reporting mAP, following the standard procedure for multi-label datasets. Finally, PASCAL VOC 2007 classification uses mean average precision (mAP), computed using the TRECVID 11-point interpolation~\cite{everingham07pascal}.\footnote{The definition of AP was changed in later versions of the benchmark.}

\paragraph{Descriptor details.} \rcnn and \dcnn build on the pre-trained VGG-M~\cite{chatfield14return} model as this performs better than other popular models such as~\cite{jia13caffe} while having a similar computational cost. This network results in 4096-dimensional \rcnn features and 512-dimensional local features for \dcnn computation. The latter are pooled into a FV representation with 64 Gaussian components, resulting in 65K-dimensional descriptors. While the \dcnn dimensionality is much higher than the 4K dimensions of \rcnn, the FV is known to be highly redundant and can be typically compressed by one order of magnitude without appreciable reduction in the classification performance~\cite{parkhi14a-compact}, so the effective dimensionality of FC- and \dcnn is likely comparable.  We verified that by PCA-reducing FV to 4096 dimensions and observing only a marginal reduction in classification performance in the PASCAL VOC object recognition task described below. In addition to VGG-M, the recent state-of-the art VGG-VD (very deep with 19 layers) model of Simonyan and Zisserman~\cite{simonyan14very} is also evaluated.

Due to the similarity between \dcnn and the dense SIFT FV descriptors used for texture recognition in~\cite{cimpoi14describing}, the latter is evaluated as well. Since SIFT descriptors are smaller (128-dimensional) than the convolutional ones (512-dimensional), a larger number of Gaussian components (256) are used to obtain FV descriptors with a comparable dimensionality. The SIFT descriptors support is $32\times 32$ pixels at the base scale.

In order to make results comparable to~\cite{cimpoi14describing}, we use the same settings whenever possible. \dcnn and D-SIFT compute features after rescaling the image by factors $2^s, s=-3,-2.5,\dots 1.5$ (but, for efficiency, discarding scales that would make the image larger than $1024^2$ pixels). Before pooling descriptors with a FV, these are usually de-correlated by using PCA. Here PCA is applied to SIFT, additionally reducing its dimension to 80, as this was empirically shown to improve the overall recognition performance. However, PCA is not applied to the convolutional features in \dcnn as in this case results were worse.

\paragraph{Learning details.} The region descriptors (\rcnn, \dcnn, and D-SIFT) are classified using 1-vs-rest Support Vector Machine (SVM) classifiers. Prior to learning, descriptors are $L^2$ normalised and the learning constant set to $C=1$. This is motivated by the fact that, after data normalisation, the exact choice of $C$ has a negligible effect on performance. Furthermore, the accuracy of the 1-vs-rest classification scheme is improved by recalibrating the SVM scores after training, by scaling the SVM weight vector and bias such that the median scores of the negative and positive training samples for each class are mapped respectively to the values $-1$ and $1$. 


\subsection{Region recognition: textures}\label{s:exp-rec}

This and the following section evaluate region recognition assuming that the ground-truth region $R$ is known~(Table~\ref{f:tex-baseline}), for example because it fills the entire image. This section focuses on textures (materials and perceptual attributes), while the next on objects and scenes.

\paragraph{Texture recognition without clutter.} This experiment evaluates the performance of \rcnn, \dcnn, D-SIFT, and their combinations in standard texture recognition benchmarks such as FMD, KTH-TIPS-2, and DTD. \rcnn is roughly equivalent to the DeCAF method used in~\cite{cimpoi14describing} for this data as regions fill images; however, while the performance of our \rcnn is similar in KTH ($\sim70\%$), it is substantially better in FMD ($60.7\% \rightarrow 70.4\%$ accuracy) and DTD ($54.8\%\rightarrow 58.7\%$). This likely is caused by the improved underlying CNN, an advantage which is more obvious in FMD and DTD that are closer to object recognition than KTH. \dcnn performs within $\pm 2\%$ in FMD and KTH but substantially better for DTD ($58.7\% \rightarrow 66.6\%$). D-SIFT is comparable in performance to \rcnn in DTD and KTH, but substantially worse ($70.4\% \rightarrow 59.2\%$) in FMD. Our conclusion is that, even when textures fill the input image as in these benchmarks, orderless pooling in \dcnn and D-SIFT can be either the same or substantially better than the pooling operated in the fully-connected layers by \rcnn.

Combining FC- and \dcnn improves performance in all datasets by $1-3\%$. While this combination is already significantly above the state-of-the-art in DTD and FMD ($+2.6\%/11.2\%$), the method of~\cite{cimpoi14describing} still outperforms these descriptors in KTH. However, replacing VGG-M with VGG-VD significantly improves the performance in all cases -- a testament to the power of deep features. In particular, the best method FC+\dcnn-VD, improves the state of the art by at least $6\%$ in all datasets. Interestingly, this is obtained by using a \emph{single} low-level feature type as FC- and \dcnn build on the same convolutional features. Adding D-SIFT results in at most $\sim 1\%$ improvement, and in some cases it slightly degrades performance.

\paragraph{Texture recognition in clutter.} The advantage of \dcnn over \rcnn is much larger when textures do not fill the image but are extracted from clutter. In OS+R (Sect.~\ref{s:data}), material recognition accuracy starts at about $46\%$ for both \rcnn and D-SIFT; however, \dcnn improves this by more than $11\%$ ($46.5\% \rightarrow 58.1\%$). The combination of FC- and \dcnn improves results further by $\sim 2\%$, but adding SIFT deteriorates performance. With the very deep CNN conclusions are similar; however, switching to VGG-VD barely affects the \rcnn performance ($46.5 \rightarrow 48.0\%$), but strongly affects the one of \dcnn ($58.1\% \rightarrow 65.1\%$). This confirms that \rcnn, while excellent in object detection, is not a very good descriptor for classifying textured regions. Results in OSA+R for texture attribute recognition (Sect.~\ref{s:data}) and in MSRC+R for semantic segmentation are analogous; it is worth noting that, when ground-truth segments are used in this experiment, the best model achieves a nearly perfect  $99.7\%$ classification rate in MSRC.

\subsection{Region recognition: objects and scenes}\label{s:exp-places}

This section shows that the \dcnn descriptor, despite its orderless nature that make it an excellent texture descriptor, excels at object and scene recognition as well. In the remainder of the section, and unless otherwise noted, region descriptors are applied to images as a whole by considering these single regions.

\paragraph{\dcnn vs \rcnn.} As seen in Table~\ref{f:tex-baseline}, in PASCAL VOC and MIT Indoor the \rcnn descriptor performs very well but in line with previous results for this class of methods~\cite{chatfield14return}. \dcnn performs similarly to \rcnn in PASCAL VOC, but substantially better ($+5\%$ for VGG-M and $+13\%$ for VGG-VD) in MIT Indoor. As further discussed below, this  is an example of the ability of \dcnn to transfer between domains better than \rcnn. The gap between \rcnn and \dcnn is the highest for the very deep VGG-VD models (68.1\% $\rightarrow$ 81.1\%), a trend also exhibited by other datasets as seen in Tab.~\ref{f:tex-baseline}. In the CUB dataset, \dcnn significantly outperforms \rcnn both whether the descriptor is computed from the whole image (CUB) or from the bird bounding box (CUB+R). In the latter case, the difference is very large ($+10-14\%$).

\paragraph{Comparison with alternative pooling methods.} \dcnn is related to the method of~\cite{gong14multi-scale}, which uses a similar underlying CNN and VLAD instead of FV for pooling. Notably, however, \dcnn results on MIT Indoor are markedly better than theirs for both VGG-M and VGG-VD ($68.8\% \rightarrow 73.5\% / 81.1\%$ resp.) and marginally better ($68.8\% \rightarrow 69.1\%$) when the same CAFFE CNN is used (Tab.~
\ref{t:places-cnn}). The key difference is that \dcnn pools convolutional features, whereas~\cite{gong14multi-scale} pools fully-connected descriptors extracted from square image patches. Thus, even without spatial information as used by~\cite{gong14multi-scale}, \dcnn is not only substantially faster, but at least as accurate. Using the same settings, that is, the same net and the same three scales, our approach results in an 8.5$\times$ speedup.

\begin{table}[t]
\centering
\begin{tabular}{|r|ccc|}
\hline
 & \multicolumn{3}{c|}{Accuracy (\%)}\\
CNN  & \rcnn & \dcnn & FC+\dcnn \\
\hline
     PLACES       & 65.0 &  67.6 &   73.1 \\
     CAFFE        & 58.6 &  69.7 &   71.6 \\
\hline
     VGG-M        & 62.5 &  74.2 &   74.4 \\
     VGG-VD       & 67.6 &  \bf{81.0} &   80.3 \\
\hline
\end{tabular}
\caption{\textbf{Accuracy of various CNNs on the MIT indoor dataset.}}\label{t:places-cnn}
\end{table}

\paragraph{Comparison with the state-of-the-art.} The best result obtained in PASCAL VOC is comparable to the current state-of-the-art set by the deep learning method of~\cite{wei14cnn:} ($85.2\% \rightarrow 85.0\%$), but using a much more straightforward pipeline. In MIT Places our best performance is also substantially superior ($+10\%$) to the current state-of-the-art using deep convolutional networks learned on the MIT Place dataset~\cite{zhou14learning} (see also below). In the CUB dataset, our best performance is a little short ($\sim 3\%$) of the state-of-the-art results of~\cite{zhang14part-based}. However, \cite{zhang14part-based} uses a category-specific part detector and corresponding part descriptor as well as a CNN fine-tuned on the CUB data; by contrast, \dcnn and \rcnn are used here as \emph{global image descriptors} which, furthermore, \emph{are the same for all the datasets considered}. Compared to the results of \cite{zhang14part-based} without part-based descriptors (but still using a part-based object detector), our global image descriptors perform substantially better ($62.1\% \rightarrow 69.1\%$).

We conclude that \dcnn is a very strong descriptor. Results are usually as good, and often significantly better, than \rcnn. In most applications, furthermore, \dcnn is many times faster as it does not require evaluating the CNN for each target image region. Finally, FC- and \dcnn can be combined outperforming the state-of-the-art in many benchmarks.

\paragraph{Domain transfer.} So far, the same underlying CNN features, trained on ImageNet's ILSVCR, were used for all datasets. Here we investigate the effect of using domain-specific features. To do so, we consider the PLACES~\cite{zhou14learning}, trained to recognize places on a dataset of about 2.5 million labeled images. \cite{zhou14learning} showed that, applied to the task of scene recognition in MIT Indoor, these features outperform similar ones trained on ILSVCR (denoted CAFFE~\cite{jia13caffe} below) -- a fact explained by the similarity of domains. Below, we repeat this experiment using FC- and \dcnn descriptors on top of VGG-M, VGG-VD, PLACES, and CAFFE.


Results are shown in Table~\ref{t:places-cnn}. The \rcnn results are in line with those reported in~\cite{zhou14learning} -- in scene recognition with \rcnn the same CNN architecture performs better if trained on the Places dataset instead of the ImageNet data ($58.6\% \rightarrow 65.0\%$ accuracy\footnote{\cite{zhou14learning} report $68.3\%$ for PLACES applied to MIT Indoor, a small difference explained by implementation details such as the fact that, for all the methods, we do not perform data augmentation by jittering.}). Nevertheless, stronger CNN architectures such as VGG-M and VGG-VD can approach and outperform PLACES even if trained on ImageNet data ($65.0\% \rightarrow 63.0\%/68.1\%$).

However, when it comes to using the filter banks with \dcnn, conclusions are very different. First, \dcnn outperforms \rcnn in all cases, with substantial gains up to $20\%$ in correspondence of a domain transfer from ImageNet to MIT Indoor. Second, \emph{the advantage of using domain-specific CNNs disappears}. In fact, the same CAFFE model that is 6.4\% worse than PLACES with \rcnn, is actually 1.5\% \emph{better} when used in \dcnn. The conclusion is that \dcnn appears to be immune, or at least substantially less sensitive, to domain shifts.

Our tentative explanation of this surprising phenomenon is that the convolutional layers are less committed to a specific dataset than the fully ones. Hence, by using those, \dcnn  tends to be a more general than \rcnn.



\subsection{Texture segmentation}\label{s:exp-seg}

The previous section considered the problem of region recognition when the region support is known at test time. This section studies the problem of recognising regions when their extent $R$ is \emph{not} known and also be estimated.

\begin{figure*}
\newcommand{\inc}[2]{%
\includegraphics[width=0.16\textwidth,frame]{figures/os-seed-01/crisp/result-#1-seg-figures/#2}}
\includegraphics[width=\textwidth]{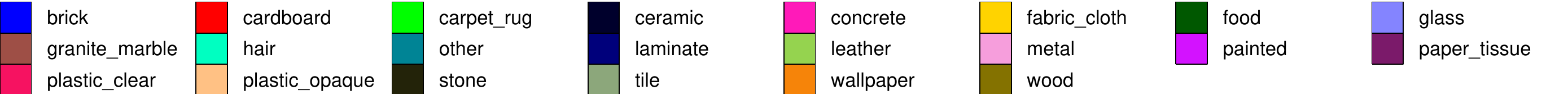}
\newcommand{\xinc}[1]{%
\inc{rcnnvd}{#1}%
\inc{rcnnvd}{#1-rcnnvd-gt}%
\inc{rcnnvd}{#1-rcnnvd-pred}%
\inc{rcnnvd}{#1-rcnnvd-err}%
\inc{dcnnvd}{#1-dcnnvd-pred}%
\inc{dcnnvd}{#1-dcnnvd-err}}\\
\null\hfill(a)\hfill\hfill(b)\hfill\hfill(c)%
\hfill\hfill(d)\hfill\hfill(e)\hfill\hfill(f)\hfill\null\\
\xinc{009626}
\xinc{008294}
\xinc{116511}
\xinc{112263}
\caption{{\bf OS material recognition results.} Example test image with material recognition and segmentation on the OS dataset. (a) original image. (b) ground truth segmentations from the OpenSurfaces repository (note that not all pixels are annotated). (c) \rcnn and crisp-region proposals segmentation results. (d) incorrectly predicted pixels (restricted to the ones annotated). (e-f) the same, but for \dcnn.}\label{f:os-examples}
\end{figure*}

\begin{figure*}
\newcommand{\inc}[2]{%
\includegraphics[width=0.16\textwidth,frame]{figures/msrc-seed-01/crisp/result-#1-seg-figures/#2}}
\includegraphics[width=\textwidth]{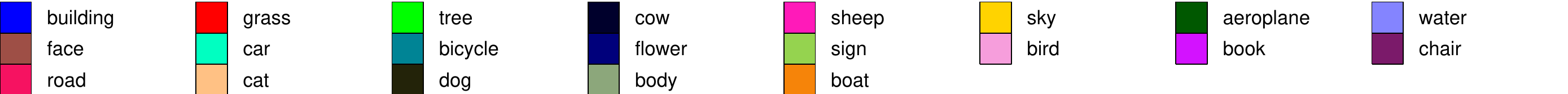}
\newcommand{\xinc}[1]{%
\inc{rcnnvd}{#1}%
\inc{rcnnvd}{#1-rcnnvd-gt}%
\inc{rcnnvd}{#1-rcnnvd-pred}%
\inc{rcnnvd}{#1-rcnnvd-err}%
\inc{dcnnvd}{#1-dcnnvd-pred}%
\inc{dcnnvd}{#1-dcnnvd-err}}\\
\null\hfill(a)\hfill\hfill(b)\hfill\hfill(c)%
\hfill\hfill(d)\hfill\hfill(e)\hfill\hfill(f)\hfill\null\\
\xinc{18_29_s}
\xinc{4_20_s}
\xinc{7_10_s}
\xinc{8_19_s}
\caption{{\bf MSRC object segmentation results.} (a) image, (b) ground-truth, (c-d) \rcnn, (d-e) \dcnn segmentation and errors.}
\label{f:msrc-examples}
\end{figure*}

The first experiment (Tab.~\ref{f:seg-table}) investigates the simplest possible scheme: combining the region descriptors of Sect.~\ref{s:exp-rec} with a general-purpose image segmentation method, namely the \emph{crisp regions} of~\cite{isola14crisp}. Two datasets are evaluated: OS for material recognition and MSRC for things \& stuff. Compared to OS+R, classifying crisp regions results in a drop of about $5\%$ points  for all descriptors. As this dataset is fairly challenging with best achievable performance is $55.4\%$, this is a satisfactory result.  But it also illustrates that there is ample space for future improvements. In MSRC, the best accuracy is $87.0\%$, just a hair above the best published result $86.5\%$~\cite{ladicky10what}. Remarkably, these algorithms not use any dataset-specific training, nor CRF-regularised semantic inference: they simply greedily classify regions as obtained from a general-purpose segmentation algorithms. Qualitative segmentation results (sampled at random) are given in Fig.~\ref{f:os-examples} and~\ref{f:msrc-examples}.

Unlike crisp regions, the proposals of~\cite{arbelaez2014multiscale} are overlapping and a typical image contains thousands of them. We propose a simple scheme to combine prediction from multiple proposals. For each proposal we set its \emph{label} to the highest scoring class, and \emph{score} to the highest score. We then sort the proposals in the increasing order of their score divided by their \emph{area} and paste them one by one. This has the effect of considering larger regions before smaller ones and more confident regions after less ones for regions of the same area. Results using \dcnn shown in Tab.~\ref{f:seg-table} in brackets (\rcnn was too slow for our experiments). The results are comparable to those using crisp regions, and we obtain $55.7\%$ accuracy on the OS dataset. Our initial attempts at schemes such as non-maximum suppression of overlapping regions that are quite successful for object segmentation~\cite{hariharan2014simultaneous} performed rather poorly. We believe this is because unlike objects, material information is fairly localized and highly irregularly shaped in an image. However, there is room for improvement by combining evidence from multiple segmentations.

\subsection{Convolutional layer analysis}\label{s:exp-filterbank}

\begin{figure}[t]
\centering
\includegraphics[width=0.41\textwidth]{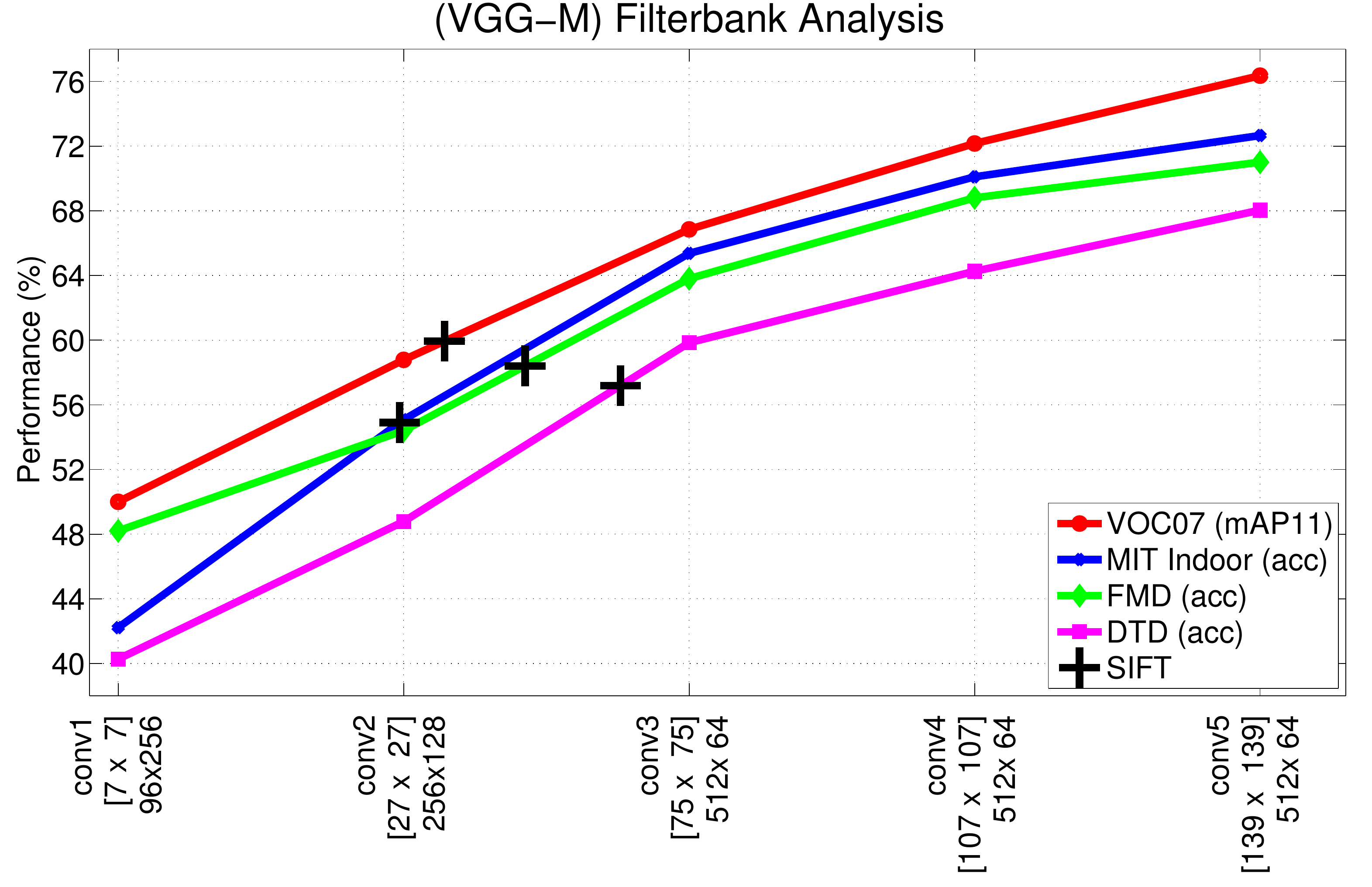}
\caption{\textbf{CNN filterbank analysis for VGG-M.} Performance of filter banks extracted from various layers network are shown on various datasets. For each layer $conv\{1,\ldots,5\}$ we show, the size of the receptive field [$N\times N$], and $FB\times D$, where $FB$ is the size of filter bank and $D$ is the dictionary size in the FV representation. The performance using SIFT is shown in black plus (+) marks. }\label{f:filterbank}
\end{figure}

We study the performance of filter banks extracted from different layers of a CNN in the \dcnn framework. We use the VGG-M network which has five convolutional layers. Results on various datasets, obtained as in Sect.~\ref{s:exp-rec} and~\ref{s:exp-places}, are shown in Fig.~\ref{f:filterbank}. In addition we also show the performance using FVs constructed from dense SIFT using a number of words such that the resulting FV is roughly the same size of \dcnn. The CNN filter banks from layer 3 and beyond significantly outperform SIFT. The performance monotonically improves from layer one to five. 

\section{Conclusions}

We have conducted a range of experiments on material and texture attribute recognition in a large dataset of textures in clutter. This benchmark was derived from OpenSurfaces, an earlier contribution of the computer graphics community, highlights the potential for collaboration between computer graphics and vision communities. We have also evaluated a number of state-of-the-art texture descriptors on these and many other benchmarks. Our main finding is that orderless pooling of convolutional neural network features is a remarkably good texture descriptor, versatile enough to dubbed as a scene and object descriptor, resulting in new state-of-the-art performance in several benchmarks.

\vspace{1em}
\paragraph{Acknowledgements} M.~Cimpoi was supported by ERC grant VisRec no. 228180 and the XRCE UAC grant. S.~Maji acknowledges a faculty startup grant from UMass, Amherst.

\vspace{5em}
\newpage
\footnotesize
\bibliographystyle{ieee}
\bibliography{bibliography}
\end{document}